# A Novel Stiffness Modulation Mechanism for Energy Efficient Variable Stiffness Actuators

Emre Sariyildiz
School of Mechanical, Materials, Mechatronic, and Biomedical Engineering,
Faculty of Engineering and Information Sciences
University of Wollongong
Wollongong, NSW, 2522, Australia
emre@uow.edu.au

*Abstract*— **This paper presents a novel mechanism that allows us to develop energy efficient variable stiffness actuators that can provide rapid and wide range stiffness modulation in practice. Although infinite-range stiffness modulation can be achieved in principal, the rigid mode of the actuator is impractical due to manufacturing limitations. The stiffness of the actuator is modulated by changing the effective length of a group of leaf springs through a novel mechanism. The nonlinear behaviours of the leaf springs enable rapid and wide-range stiffness modulation, which can provide many benefits in advanced robotic applications such as human-robot collaboration and safe physical robot environment interaction. The leaf springs also allow us to develop energy efficient variable stiffness actuators. The servo system that modulates the mechanical stiffness does not drain power to keep the stiffness of the actuator fixed at equilibrium positions. Therefore, the proposed variable stiffness actuator is more energy efficient than the conventional antagonistic actuators that constantly drain power for stiffness modulation. When the stiffness of the actuator is changed at equilibrium positions, the power drain of the stiffness modulation motor due to inertial and frictional disturbances is negligible. The disturbances of the stiffness modulation motor exerted by the novel mechanism increases as the deflection angle of the output link increases. Nevertheless, the power drain by the stiffness modulation motor is always bounded at non-equilibrium positions. A variable stiffness actuator can be easily developed by placing the novel stiffness modulation mechanism onto a rigid actuator's output. Simulations and experiments are presented to verify the proposed novel stiffness modulation mechanism.**

## I. INTRODUCTION

The stiff and non-backdrivable mechanical structures of conventional rigid actuators enable us to perform high-precision position control tasks using simple motion controllers such as Proportional-Integral-Derivative (PID) controllers [1–6]. Although the conventional rigid actuators are very useful for many industrial tasks (e.g., pick and place) that merely require high-precision positioning, they lead to severe safety problems when robots interact with unknown and dynamic environments [6–9]. To tackle the safety problem by controlling the interaction forces between robots and environment, many implicit and explicit force controllers have been proposed in the literature [10–14]. Nevertheless, the poor stability and low-performance of force control systems still remain a great challenge when robots interact with unstructured environments [14–16]. Moreover, the trade-off between the torque density and mechanical backdrivability of conventional actuators present many practical limitations in robot environment interaction [9].

To meet the requirements of many advanced robotic systems that interact with open and dynamic environments (e.g., cobots working alongside human workers), mechanically compliant actuation systems have received increasing attention in the last decades [17–19]. For example, while many soft actuators have been proposed by printing soft materials with different 3D printing techniques such as Fused Deposition Modelling (FDM) and Stereolithography (SLA) [17], series elastic actuators have been developed by placing a compliant mechanical element, such as torsional spring, between the rigid actuator and link [3, 18–20]. The mechanical compliance can simply improve safety by absorbing impact forces in collision [21, 22]. Moreover, the mechanically backdrivable structure of compliant actuators allows us to use simple motion controllers (e.g., joint space position controllers) for robot environment interaction applications [2, 18, 19]. Despite many benefits in safe robot-environment interaction, the compliant actuators present several challenges in practice. For example, the output force of soft robotic systems are generally limited compared to conventional rigid actuators [17, 21, 23]. Although the output force can be increased using different compliant actuators such as series elastic actuators, the compliant mechanical component between the rigid actuator and link leads to severe bandwidth limitations and challenging position control problems [24–26]. Thus, robotic systems developed by compliant actuators present low-performance in many practical applications such as vibration in pick and place tasks. To tackle this problem, advanced motion controllers should be employed in the position control tasks of compliant actuators [24–27].

A simple yet efficient solution for the position and physical interaction problems of robotic system can be obtained by using an actuator that can adjust its compliance mechanically [28–30]. While the position control tasks can be performed using simple motion controller in the stiff mode of the actuator, the soft mode of the actuator can improve safety in physical interaction tasks [19, 31]. Such actuators are generally called variable impedance or variable stiffness actuators in the literature [28]. Antagonistic actuators are one of the most widely used variable stiffness actuators since 1980s [32–34]. In the conventional antagonistic actuation inspired by mammalian anatomy, equilibrium position is controlled by rotating two motors, which are connected to the output link through nonlinear springs, in the same direction [33]. The stiffness of the actuator is modulated by controlling the tensions of the nonlinear springs [33]. This leads to high-energy consumption as the motors constantly drain power due to the nonlinear springs' disturbances [30, 31]. In addition, the coupled motion and stiffness modulation control problem, limited output torque and limited potential energy capacity are the other drawbacks of the conventional antagonistic actuation system [30, 31]. To solve these problems, different antagonistic and non-antagonistic variable stiffness actuators have been proposed in the literature [31]. While the antagonistic actuators generally suffered from similar drawbacks, some non-antagonistic actuators showed promising results, e.g., decoupled motion control and stiffness modulation problem in [35, 36], energy efficient variable stiffness actuation in [37, 38], and infinite-range stiffness modulation in [39]. Nevertheless, to the best of our knowledge, no variable stiffness actuator can meet the requirements of many advanced robotic applications [30, 31].

To this end, this paper proposes a novel stiffness modulation mechanism that can provide rapid stiffness modulation in a wide stiffness range, from a very low stiffness value (i.e., soft mode) to infinite stiffness (i.e., rigid mode) in principal. A variable stiffness actuator can be easily developed by placing the novel stiffness modulation mechanism to the output of a conventional rigid actuator. The stiffness of the actuator can be kept fixed at equilibrium positions without draining power by the stiffness modulation motor. When the stiffness of the actuator is altered, the stiffness modulation motor is disturbed by inertial and frictional disturbances only. The energy consumption of stiffness modulation is therefore negligible at equilibrium positions. The disturbances exerted on the stiffness modulation motor increases as the deflection of the output link gets larger. However, the power consumption of stiffness modulation is always bounded. The unique features of the proposed variable stiffness actuation mechanism can provide several benefits in next generation robotic applications. For example, a cobot can safely work alongside human workers in the soft mode of the actuator while the rigid mode enables high-precision position control tasks repeatedly. Moreover, mobile autonomous robots can perform more tasks in a single battery run time by using energy efficient variable stiffness actuators. Simulations and experiments are presented to verify the proposed novel stiffness modulation mechanism.

The rest of the paper is organised as follows. Section II presents a new variable stiffness actuator developed by using the novel stiffness modulation mechanism. Section III proposes simple yet efficient models for the variable stiffness actuator and novel stiffness modulation mechanism. Section IV presents simulation and experimental results to verify the proposed novel actuation system. The paper ends with conclusion in Section V.

## II. A Variable Stiffness Actuator Developed using the Novel Stiffness Modulation Mechanism

### A. *Variable Stiffness Actuator*

Figure 1 illustrates a new variable stiffness actuator that is developed by integrating the novel stiffness modulation mechanism onto a conventional rigid actuator. It comprises a servo motor (motor 1) that is used to regulate the equilibrium position of the output link, second servo motor (motor 2) that modulates the stiffness of the actuator, and the novel stiffness modulation mechanism.

As shown in Fig. 1, the proposed variable stiffness actuator has a modular structure. It can be developed by combining a conventional rigid actuator and the novel stiffness modulation mechanism. This modular mechanical structure provides significant benefits in design and manufacturing. For example, the torque density of the actuator can be easily adjusted for different robotic applications by modifying the first motor (i.e., motor 1 in Fig. 1) and gearbox combination of the conventional rigid actuator. This also allows us to build compact variable stiffness actuators. The stiffness range of the actuator can be tuned by changing the material, design and the number of the leaf springs in the stiffness modulation mechanism. Therefore, softer and stiffer actuators can be easily developed using the proposed novel actuation system. The stiffness modulation characteristics of the actuator, such as the stiffness modulation speed and power consumption, can also be tuned using different motor and ball screw combinations in the design of the novel stiffness modulation mechanism.

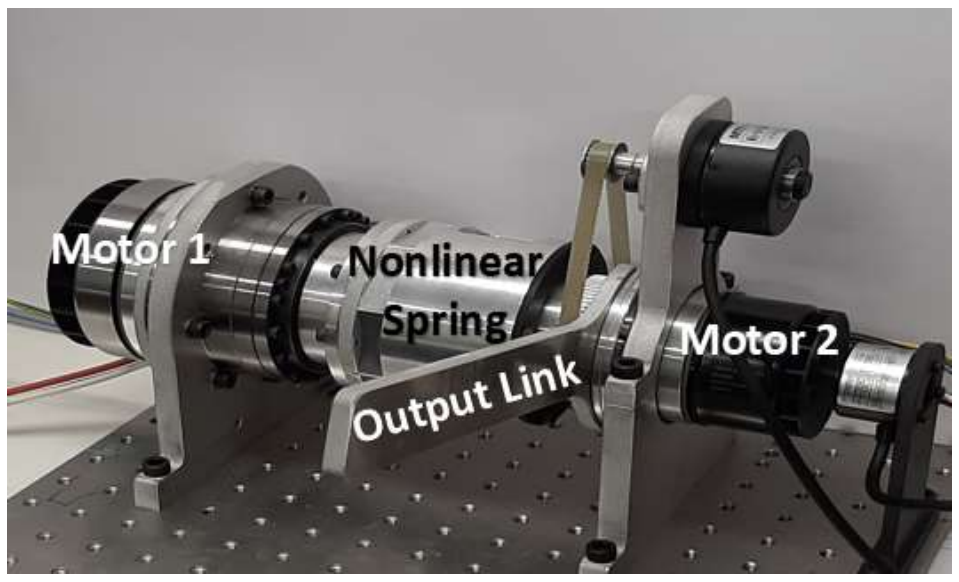

Fig. 1: A variable stiffness actuator developed by using the novel stiffness modulation mechanism.

The decoupled motion control and stiffness modulation characteristic is the other important feature of the proposed variable stiffness actuator. While the first motor independently regulates the equilibrium positions of the output link, the stiffness of the actuator can be directly controlled by the second motor. The decoupled characteristic of the variable stiffness

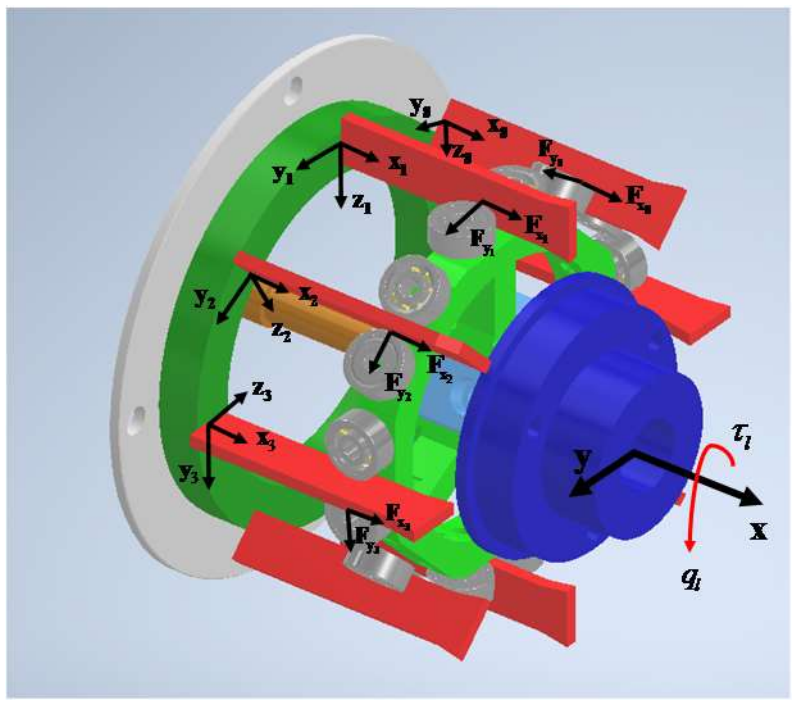

a) CAD design of the stiffness modulation mechanism.

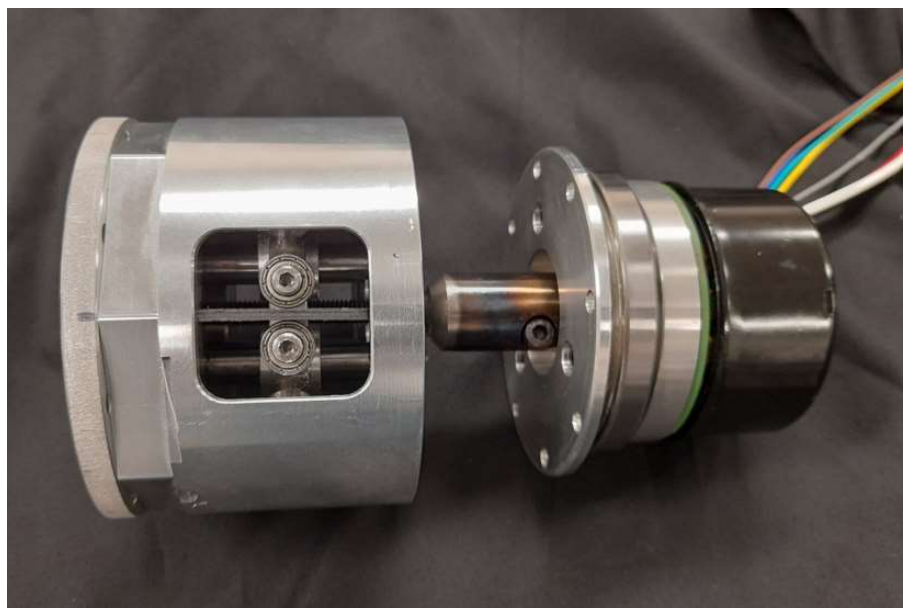

b) First prototype of the stiffness modulation mechanism.

Fig. 2: CAD design and first prototype of the stiffness modulation mechanism.

actuator allows us to perform motion control and stiffness modulation tasks using simple controllers. The deflection of the link and the stiffness of the mechanism change the actuator's output torque independently, so the output torque can be controlled by both of the servo systems.

## B. Novel Stiffness Modulation Mechanism

The CAD model and first prototype of the novel stiffness modulation mechanism are illustrated in Fig. 2. As shown in this figure, the mechanism is developed using eight leaf springs connected in parallel and a linear actuation system comprising a ball screw and a dc motor, i.e., motor 2 in Fig. 1. Rollers are placed on the leaf springs to reduce the friction of the linear actuation system.

The stiffness of the mechanism is modulated by changing the effective lengths of the leaf springs via the linear actuation system. The positions of the rollers represent the effective lengths of the leaf springs where the concentrated forces are exerted on the springs due to the output torque. As the rollers move towards the free end, the stiffness of the actuator decreases. The stiffness of the actuator becomes the largest (theoretically infinite) when the rollers are at the fixed end. The nonlinear characteristics of the leaf springs enable us to perform a wide-range stiffness modulation rapidly.

In addition, the proposed variable stiffness actuation mechanism enables energy efficient stiffness modulation. When the actuator is at equilibrium positions (i.e., the deflection angle of the output link is zero), the novel stiffness modulation mechanism does not consume power to keep the stiffness of the actuator constant. The power consumption of stiffness modulation is also negligible at equilibrium positions because only inertial and frictional disturbances exert on the stiffness modulation motor. In non-equilibrium positions, disturbances due to stiffness modulation increase, leading to higher power consumption. However, the power consumption of the proposed actuation system is always bounded.

# III. Model of a Variable Stiffness Actuator Developed by the Novel Stiffness Modulation Mechanism

## A. Variable Stiffness Actuator

The dynamic model of the variable stiffness actuator can be derived using Fig. 3 as follows:

$$\begin{aligned} J_{m1}\ddot{q}_{m1} + b_{m1}\dot{q}_{m1} &= \tau_{m1} - \tau_s \\ J_{m2}\ddot{q}_{m2} + b_{m2}\dot{q}_{m2} &= \tau_{m2} - \tau_s^d \\ J_l\ddot{q}_l + b_l\dot{q}_l &= \tau_s - \tau_l \end{aligned} \quad (1)$$

where $J_{m1}$, $J_{m2}$ and $J_l$ are the inertiæ of the motor 1 that controls the equilibrium position of the actuator, motor 2 that controls the stiffness modulation, and output link, respectively; $b_{m1}$, $b_{m2}$ and $b_l$ are the viscous friction coefficients of the motor 1, motor 2, and output link, respectively; $q_{m1}$, $q_{m2}$ and $q_l$ are the position states of the motor 1, motor 2, and output link, respectively; $\dot{q}_\bullet$ and $\ddot{q}_\bullet$ are the first and second derivatives of $q_\bullet$, i.e., velocity and acceleration states of the actuator; $\tau_{m1}$ and $\tau_{m2}$ are the thrust torques of the motor 1 and motor 2, respectively; $\tau_s$ is the torque exerted by the nonlinear spring of the variable stiffness actuation mechanism; and $\tau_s^d$ is the disturbance torque exerted by the nonlinear spring on motor 2 [27]. In Fig. 3, $k$ represents the stiffness of the variable stiffness actuator. Similar to $\tau_s$ and $\tau_s^d$, the stiffness of the actuator $k$ can be represented as a nonlinear function of the position states of the novel actuation system, i.e., $\tau_s(q_{m1}, q_{m2}, q_l)$, $\tau_s^d(q_{m1}, q_{m2}, q_l)$ and $k(q_{m1}, q_{m2}, q_l)$. The nonlinear functions of the stiffness and spring torques are derived in Section III.B.

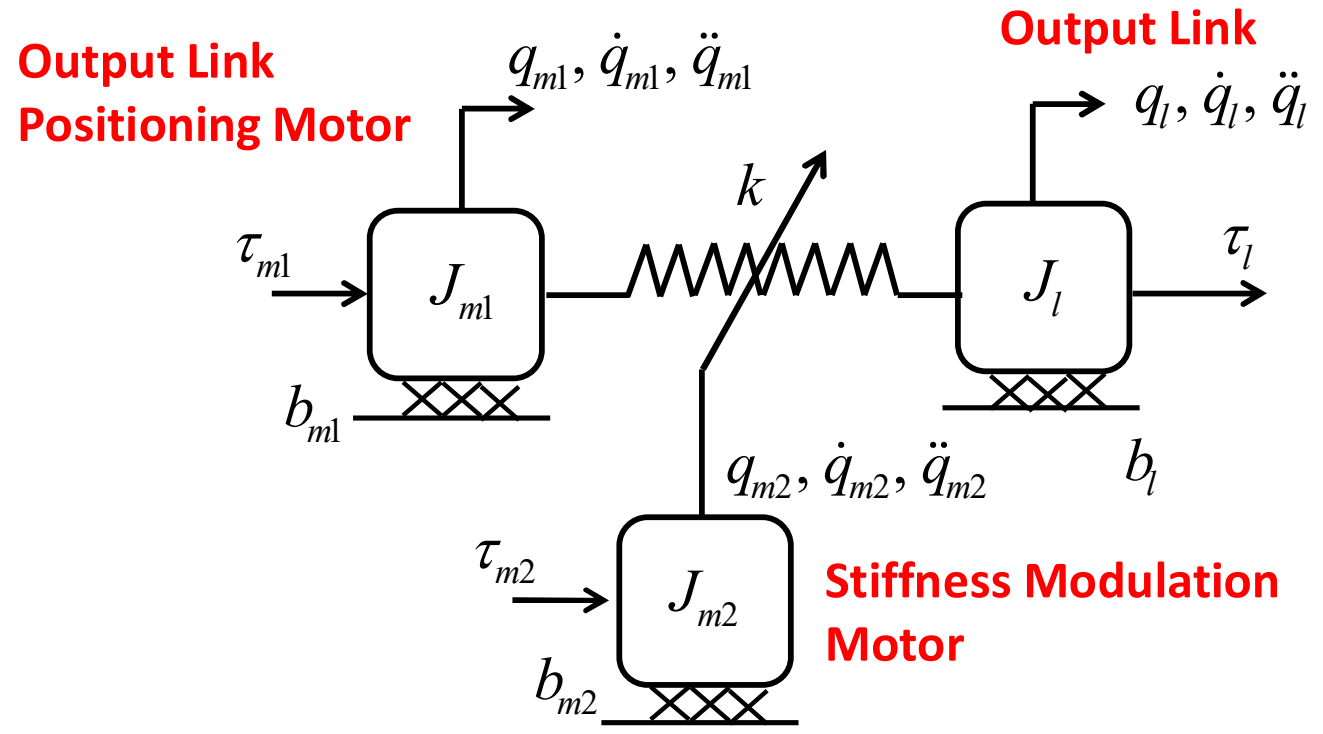

Fig. 3: A dynamic model for the variable stiffness actuator developed by the novel stiffness modulation mechanism.

By using the analogy of a mass-spring-damper system, Eq. (1) represents a simple yet efficient dynamic model for the variable stiffness actuators developed by using the novel stiffness modulation mechanism. To find the torques exerted on the motors and output link, the model of the nonlinear spring, i.e., the nonlinear functions $\tau_s$ and $\tau_s^d$ in Eq. (1), should be derived. Moreover, the model of the nonlinear spring's stiffness $k$ should be derived to clarify the stiffness modulation characteristics of the proposed variable stiffness actuation system.

## B. Novel Stiffness Modulation Mechanism

Let us consider the first leaf spring only, i.e., the stiffness modulation mechanism is designed using a single leaf spring. When the torsional motion is neglected, the 2-dimensional model of the first leaf spring can be represented in Fig. 4. In this figure, the following apply:

| | |
|---|---|
| $x_1$ and $y_1$ | positions along $\mathbf{x_1}$ and $\mathbf{y_1}$ axes; |
| $S_{x_1}$ | arc length of the beam at $x_1$; |
| $\phi_{x_1}$ | slope of the beam at $x_1$; |
| $\delta_{x_1}$ and $\delta_{y_1}$ | beam deflection at $x_1$ along $\mathbf{x_1}$ and $\mathbf{y_1}$ axes; |
| $\delta_{x_r}$ and $\delta_{y_r}$ | beam deflection at $x_r$ along $\mathbf{x_1}$ and $\mathbf{y_1}$ axes; |
| $L$ | length of the beam; |
| $\boldsymbol{F}$ | forces acting on the beam; and |
| $\phi_{x_1} = \phi_{x_r}$ | slope of the beam at $x_r$. |

where $x_1$ represents a random point on the beam and $x_r$ represents the location of the rollers on the $\mathbf{x_1}$-axis [30, 31, 40].

Using Euler-Bernoulli beam theory and Fig. 4, the relation between the applied force $F_{y_1}$ and deflections along $\mathbf{x_1}$- and $\mathbf{y_1}$-

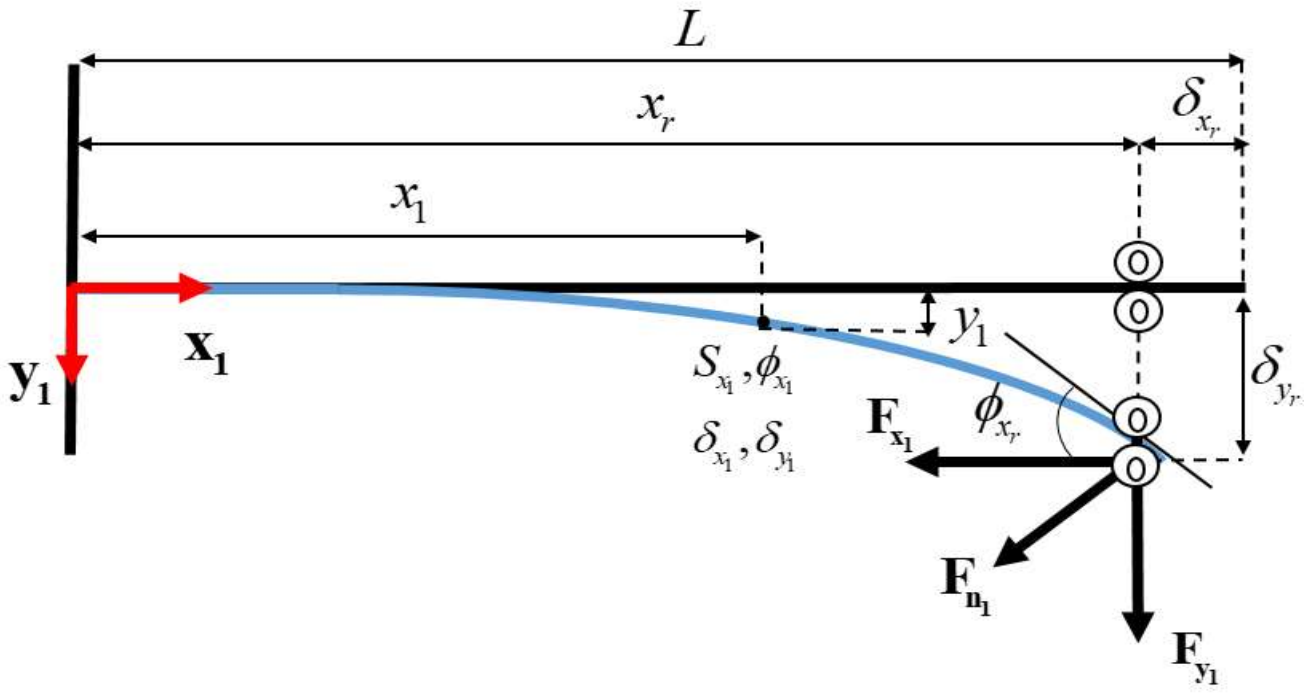

Fig. 4: 2D deflection model of the first leaf spring.

axes can be derived as follows:

$$S_{x_1} = \sqrt{\frac{EI}{2F_{y_1}}} \int_0^{\phi_{x_1}} \frac{d\phi}{\sqrt{\sin(\phi_{x_r}) - \sin(\phi)}} \tag{2}$$

$$x_1 = \sqrt{\frac{2EI}{F_{y_1}}} \int_0^{\phi_{x_1}} \frac{\cos(\phi)\, d\phi}{\sqrt{\sin(\phi_{x_r}) - \sin(\phi)}} \tag{3}$$

$$y_1 = \sqrt{\frac{2EI}{F_{y_1}}} \int_0^{\phi_{x_1}} \frac{\sin(\phi)\, d\phi}{\sqrt{\sin(\phi_{x_r}) - \sin(\phi)}} \tag{4}$$

where $E$ is Young's modulus, and $I$ is the moment of inertia of the beam cross section about the neutral axis [30, 31, 40].

Although Eq. (2) – Eq. (4) can accurately calculate the deflection of the leaf spring using numerical solution methods such as Adomian decomposition method, they are usually computationally demanding so they may not be useful for real-time motion control applications. Moreover, they are not useful to understand the stiffness modulation characteristics of the variable stiffness actuation mechanism.

To derive a more practical model for the novel stiffness modulation mechanism, let us assume that the deflection of the leaf spring is small. This assumption allows us to obtain the following spring torque and stiffness equations using the conventional simple beam theory.

$$\tau_s(q_{m1}, q_{m2}, q_l) = \frac{6EI}{x_r^3} r^2 \sin\left(\frac{\Delta_{q_l}}{2}\right) \tag{5}$$

$$k(q_{m1}, q_{m2}, q_l) = \frac{3EI}{x_r^3} r^2 \cos\left(\frac{\Delta_{q_l}}{2}\right) \tag{6}$$

where $\Delta_{q_l} = q_l - q_{m1}$ represents the deflection angle of the output link; $x_r = \frac{\ell}{2\pi} q_{m2}$ represents the position of the rollers, i.e., the position of the force applied to the leaf spring, in which $\ell$ represents the lead of the ball screw [31, 40].

Using simple beam theory, a simple yet efficient model given in Eq. (5) and Eq. (6) can be achieved for the novel stiffness actuation mechanism. As shown in Eq. (5), the output torque $\tau_s$ gets higher as the deflection angle of the output link increases. Moreover, the output torque is regulated by adjusting the position of the rollers on the beam through the motion control of the second motor. As the position of the roller gets closer to the leaf spring's fixed end, i.e., $x_r$ gets smaller, the output torque of the actuator increases. While the deflection of the output link directly changes the reaction torque, the position of the second motor implicitly controls the output torque through stiffness modulation as shown in Eq. (6). This equation shows that the stiffness of the motor changes by the deflection of the output link and the position of the second motor. The stiffness of the variable stiffness actuator can be independently controlled using the latter control parameter. As shown in Eq. (6), the nonlinear stiffness modulation characteristic allows us to modulate the stiffness of actuators rapidly. This can provide significant benefits in practical motion control applications, such as pedal locomotion and human-robot collaboration [31].

Using small deflection assumption, the disturbance torque on the second motor due to stiffness modulation can be derived as follows:

$$\tau_s^d = \frac{6EI}{x_r^3} r^2 \sin\left(\frac{\Delta_{q_l}}{2}\right) \tan(\varphi) \tag{7}$$

where $\varphi = \frac{F_y L^2}{2EI}$ [31, 40].

Equation (7) shows that when the stiffness of the actuator is kept fixed at equilibrium positions, i.e., $\Delta_{q_l} = 0$ and $\dot{q}_{m2} = 0$, the novel variable stiffness actuation mechanism does not consume power. Therefore, it provides more energy efficient stiffness modulation characteristic than conventional antagonistic variable stiffness actuators. When the stiffness of the actuator is modulated in equilibrium positions, i.e., $\Delta_{q_l} = 0$ and $\dot{q}_{m2} \neq 0$, the power consumption is due to the disturbances of the linear actuation mechanism, e.g., the frictions of the ball screw mechanism and rollers on the leaf springs. Therefore, the power consumption of the novel stiffness modulation mechanism is negligible at equilibrium positions, i.e., $\Delta_{q_l} = 0$.

As shown in the Eq. (7), the disturbance torque on the stiffness modulation motor becomes higher as the deflection angle of the output link increases at non-equilibrium positions. The power consumption due to stiffness modulation is bounded at non-equilibrium positions.

When the other leaf springs are considered, the model of the novel variable stiffness actuation mechanism is obtained as follows:

$$\tau_s(q_{m2}, q_l) = \frac{48EI}{x_r^3} r^2 \sin\left(\frac{\Delta_{q_l}}{2}\right) \tag{8}$$

$$k(q_{m2}, q_l) = \frac{24EI}{x_r^3} r^2 \cos\left(\frac{\Delta_{q_l}}{2}\right) \tag{9}$$

$$\tau_s^d = \frac{48EI}{x_r^3} r^2 \sin\left(\frac{\Delta_{q_l}}{2}\right) \tan(\varphi) \tag{10}$$

## IV. Simulations and Experiments

### A. Simulations

Figure 5a illustrates the relation between the output torque and link's deflection when the stiffness of the actuator is set at

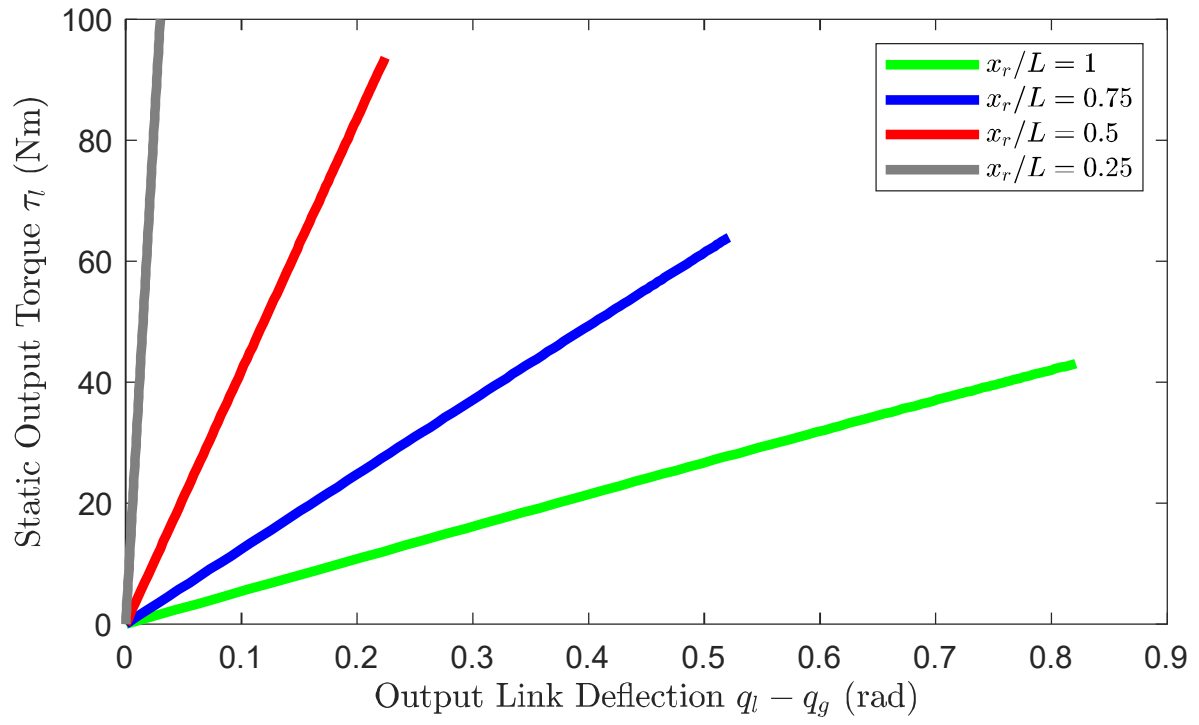

a) Static Output Torque vs Output Link Deflection.

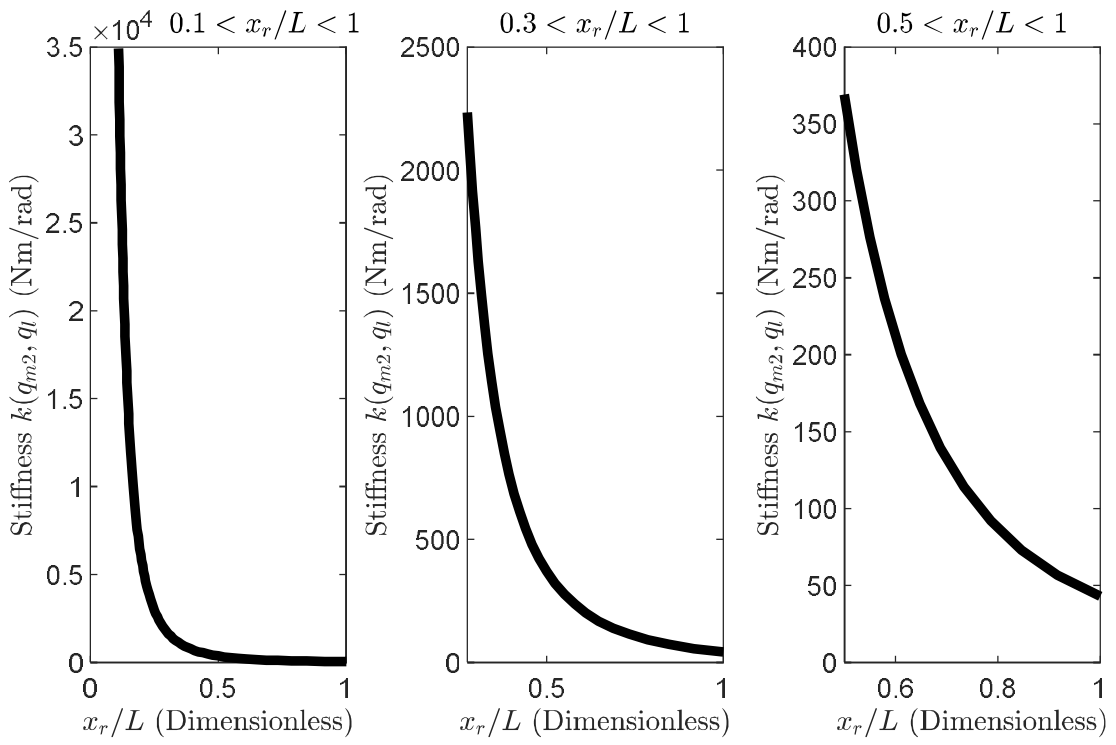

b) Stiffness modulation characteristics.

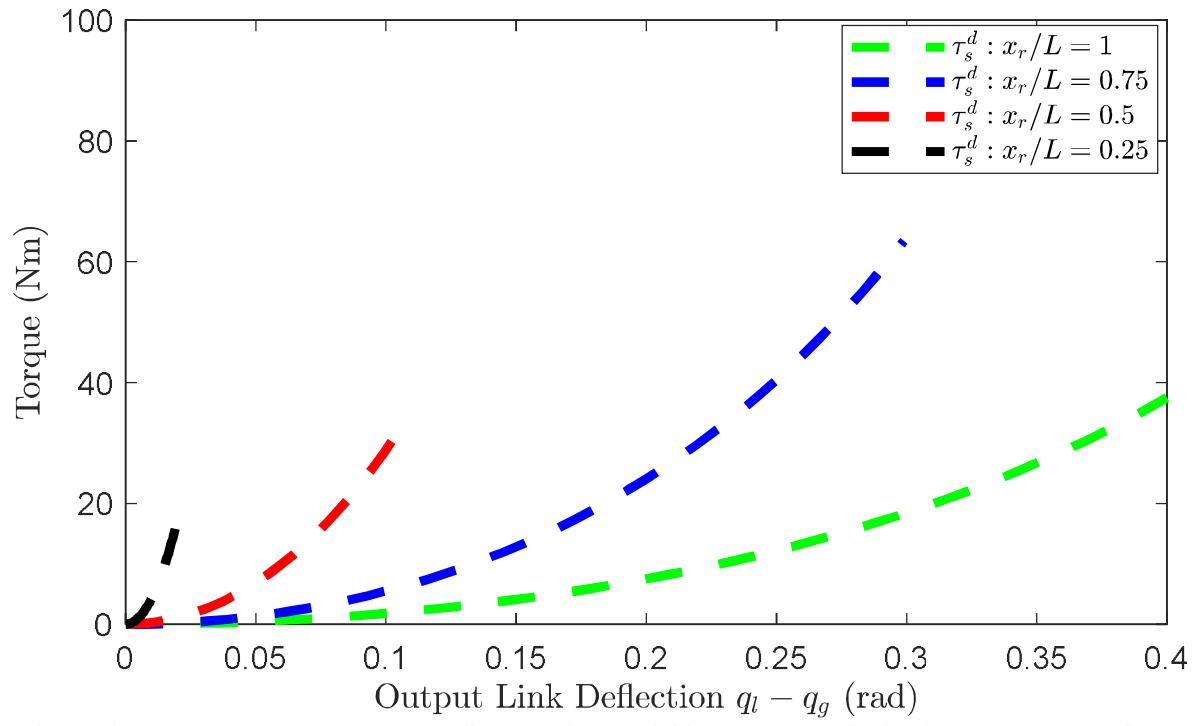

c) Disturbance torque exerted on the stiffness modulation motor due to the deflection of the output link.

Fig. 5: Static output torque, stiffness modulation characteristic and disturbance torque of the variable stiffness actuation mechanism.

different values. Since the actuator is stiffer when the rollers are closer to the fixed end of the leaf springs, the output torque of the actuator is higher for the small deflections of the output link. For example, the grey and green curves illustrate the stiffest and softest motions of the variable stiffness actuator in this figure, respectively.

Figure 5b illustrates the stiffness modulation characteristic of the variable stiffness actuation mechanism with respect to the effective length of the leaf springs (i.e., roller's position). As the rollers move towards the fixed end, the stiffness of the actuator increases exponentially, thus enabling a wide-range of stiffness change in a rapid manner. The figure shows that the stiffness of the actuator goes to infinite, i.e., the actuator becomes rigid, as $x_r$ goes to zero. Nevertheless, $x_r$ can never become zero due to practical limitations in manufacturing.

Figure 5c illustrates the disturbance torques exerted on the stiffness modulation motor. As shown in this figure, the disturbance torque is zero for all stiffness values of the actuator when the deflection angle of the output link is zero. Therefore, the power consumption of the variable stiffness actuation system is zero or negligible when the output link is at or near an equilibrium position. As the deflection of the output link increases, the disturbance torque exerted on the motor 2 becomes larger, thus leading to higher power consumption in stiffness modulation. The disturbance torque is bounded for all stiffness and output link deflection as shown in Fig. 5c.

## B. *Experiments*

Figure 6a illustrates the output torque of the variable stiffness actuator in softer and stiffer modes when the deflection of the output link is controlled using a position controller. As shown in this figure, the interaction force is smaller for the same deflection angle of the output link when the actuator is in the softer mode. This feature can be used to perform safer physical robot-environment interaction applications.

Figure 6b illustrates the stiffness modulation characteristic of the variable stiffness actuator at an equilibrium position. When the stiffness of the actuator is fixed, the power drained by the motor 2 is zero because there is no disturbance acting on the motor. When the stiffness of the actuator is altered from its minimum value to the maximum one, the power drain due to disturbances such as frictions is negligible as shown in the figure. Therefore, energy efficient stiffness modulation can be achieved using the proposed variable stiffness actuator.

## V. CONCLUSION

In this paper, a new variable stiffness actuator is presented. The actuator is developed using a novel stiffness modulation mechanism which enables rapid and energy efficient stiffness modulation from very low stiffness values to infinite-stiffness in

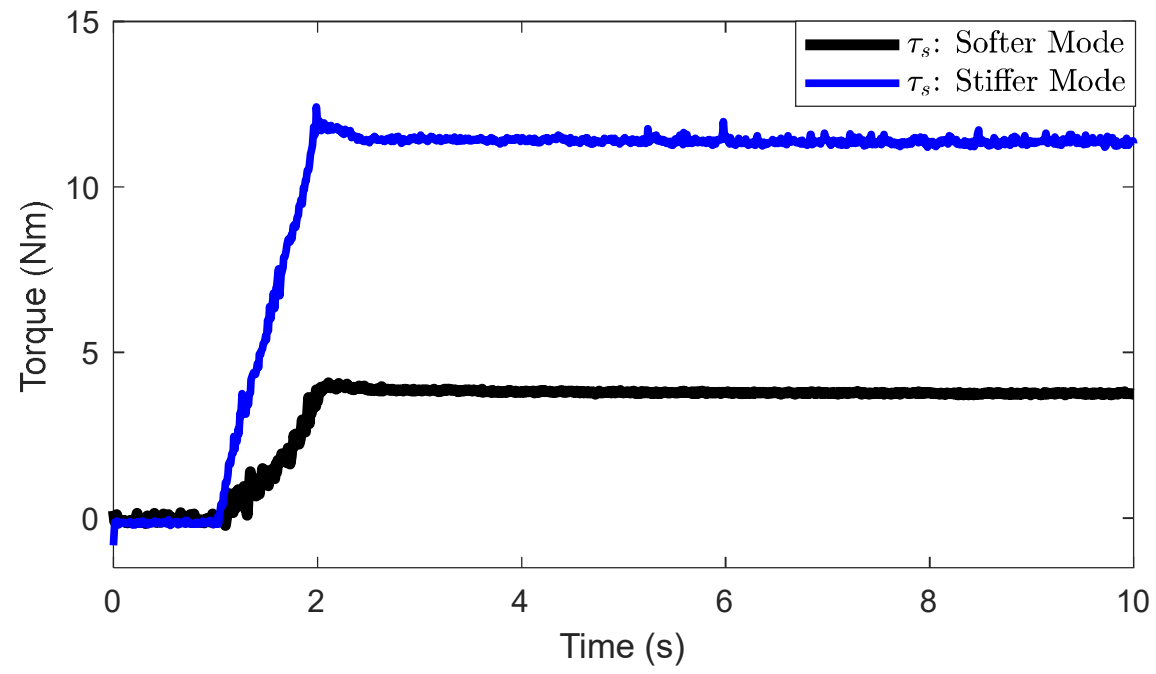

a) Output torques when the deflection of the link is controlled.

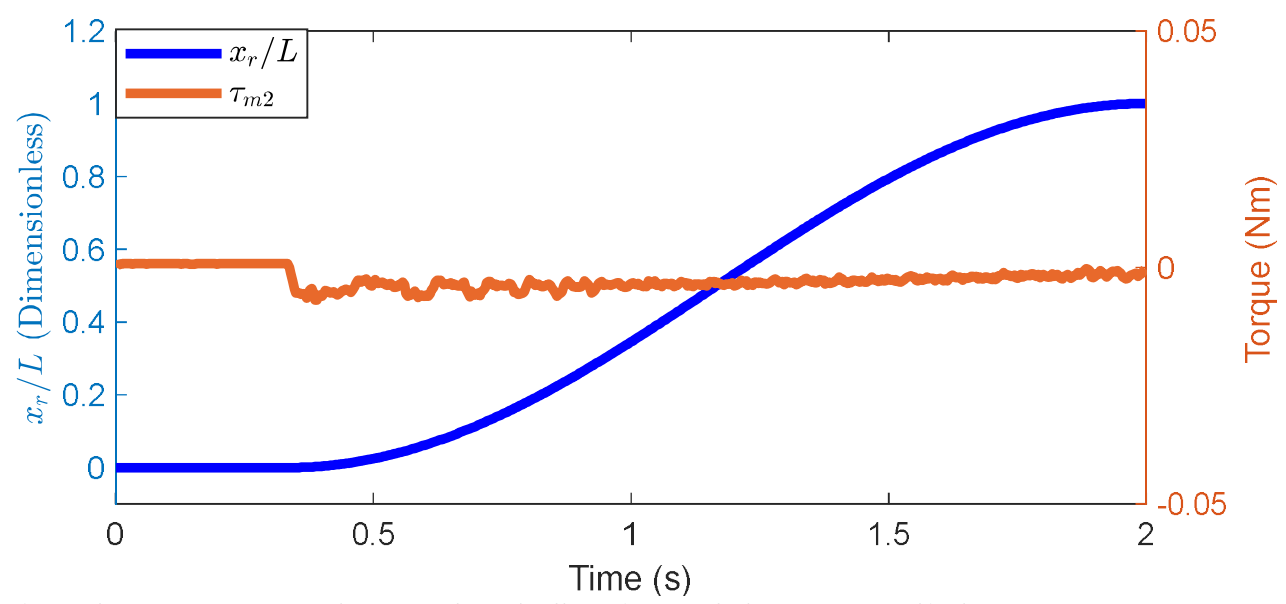

b) Disturbance torque due to the deflection of the output link.

Fig. 6: The output torque and stiffness modulation of the variable stiffness actuator.

principal. However, only very large stiffness values can be achieved due to practical manufacturing limitations. The unique features of the novel stiffness modulation mechanism can provide several benefits in advanced robotic applications. While autonomous robots can perform tasks longer using energy efficient actuators, the infinite range stiffness modulation allows us to integrate the proposed novel mechanism to various robotic systems such as exoskeletons and cobots. Moreover, the rapid stiffness modulation can enable us to improve the performance of pedal locomotion using the stiffer and softer modes of the actuator in swing and stance phases effectively.

The proposed variable stiffness actuator can be easily developed by placing the novel stiffness modulation mechanism onto a conventional rigid actuator. This allows us to easily develop actuators with different torque density and working ranges [41]. However, more experiments should be conducted to clarify the performance of the novel stiffness modulation mechanism in position and force control applications. Moreover, it should be integrated into robotic systems to better understand how the novel stiffness modulation mechanism can improve robotic applications.